\documentclass[11pt,a4paper]{article}
\usepackage{times,latexsym}
\usepackage{url}
\usepackage[T1]{fontenc}

\usepackage[acceptedWithA]{tacl2021v1} 

\usepackage{graphicx}
\usepackage{subcaption}
\usepackage{booktabs}
\usepackage{makecell}
\usepackage{multirow}
\usepackage{enumitem}
\usepackage{amsmath}
\usepackage{xcolor}
\usepackage{amssymb}
\usepackage{bm}
\usepackage{tabularx}
\usepackage[table]{xcolor}
\usepackage{float}

\usepackage{xspace,mfirstuc,tabulary}

\newif\iftaclinstructions
\taclinstructionsfalse 
\iftaclinstructions
\renewcommand{\confidential}{}
\renewcommand{\anonsubtext}{(No author info supplied here, for consistency with
TACL-submission anonymization requirements)}
\newcommand{\instr}
\fi

\iftaclpubformat 

\else

\fi

\title{Interpreting Narrative Mechanisms in Language Models}

\author{
  David Yiyang Liu\thanks{Correspondence: \texttt{david.liu2@unsw.edu.au}}
  \\
  UNSW Sydney
  \And
  Aditya Joshi
  \\
  UNSW Sydney
  \And
  Paul Dawson
  \\
  UNSW Sydney
  \And
  Andrew Piper
  \\
  McGill University
}

\date{}

\begin{document}
\pagestyle{plain} 
\maketitle
\thispagestyle{plain} 
\begin{abstract} 
Narratology is the field of study concerned with theorising properties of narrative communication. We propose extending its objects of study to include language models (LMs) and present a mechanistic study of how LMs represent \emph{narrativity}, the degree to which a text exhibits narrative properties. Across GPT-2-Small, Gemma-2-2B, and Llama-3.1-8B, we identify attention heads, MLP neurons, and SAE features whose activations correlate with narrativity. Ablating these components disproportionately increases loss on high-narrativity texts and reduces the narrativity of generations, but benchmark results are mixed. Neuronpedia explanations of identified SAE features carry signals that correspond with theoretical cognitive narrative dimensions. We propose new research directions in which language models can be studied as narrative models, with their computational mechanisms interpretable as theoretical components, and investigate how they contribute to downstream behaviours.
\end{abstract}

\section{Introduction}

\begin{figure}[t]
    \centering
    \includegraphics[width=\linewidth]{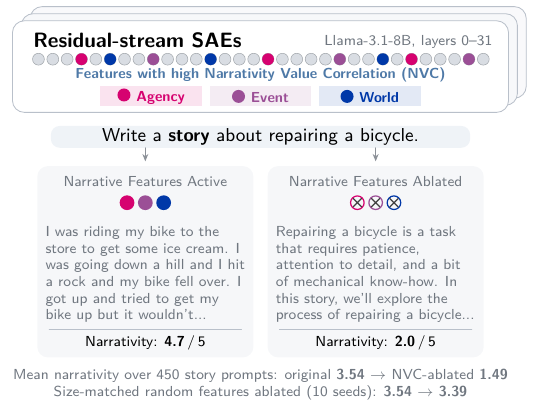}
    \caption{We interpret narrative features along three dimensions: agency, event sequencing, and world building.} 
    \label{fig:first}
\end{figure}

When a language model (LM) is prompted to write a story, its output relies upon a set of learned internal representations of \emph{what a story is}. In this paper, we investigate how an LM's internal components (attention heads, neurons and residual streams) can provide access for mechanistic interpretation of the LM's \textit{implicit theory of narrative}: a set of expectations about how stories are structured, and whose contributions to output behaviour can be identified and ablated (Figure \ref{fig:first}). In doing so, we empirically study the relationship between narrative concepts and LM mechanisms, proposing LMs as narrative models: new objects of study and experiment for narrative understanding.



We bring together narratology and mechanistic interpretability to investigate which properties of narrative LMs learn to represent, what capabilities these representations support, and what they reveal about LMs as models of narrative. Narratology theorises concepts for defining narrative \cite{genette1980narrative, herman2009basic, piper-etal-2021-narrative} but whether these concepts correspond to consequential distinctions inside an LM remains an open question. Mechanistic interpretability provides a means of investigating this correspondence: identifying internal components associated with narrative properties and ablating them to observe the consequences \citep{zhang2024towards}. 


Our work extends the growing body of LM research which examines narrative properties, such as the distribution of narrative features across sources of pre-training data \citep{johnson2026characterizing}, the presence of recurrent or homogeneous story patterns in generated outputs \citep{tian-etal-2024-large-language,hamilton2026elias} and the probing of narrative coherence represented in hidden states \citep{de2026mary}. Existing studies leave open how an LM's internal components may \emph{causally contribute} to narrative representation and downstream behaviours. Motivated by this gap, we ask:
\textbf{(1)} What representations of narrative are used by an LM during inference?
\textbf{(2)} Are those representations compatible with formal dimensions from cognitive narratology?
\textbf{(3)} Does an LM use these narrative mechanisms for `intelligent activity' \citep{herman2013storytelling}, such as common-sense reasoning and confabulation?  

These questions form new research directions which we begin to explore by studying components at three locations in the Transformer architecture (Figure \ref{fig:demo}): attention heads, neurons and latent residual stream features discovered by sparse autoencoders (SAEs) \citep{huben2024sparse}. We identify and ablate components associated with the concept of \textit{narrativity}: the degree to which a text exhibits distinct narrative properties \cite{giora1994degrees, pianzola2018looking}. 

Applying the principles of cognitive narratology \citep{herman2009basic}, we operationalise narrativity through three dimensions: \textit{agency}, the prominence and continuity of agents experiencing events; \textit{event sequencing}, the temporal organisation of those events; and \textit{world building}, the construction of a coherent, experienceable world. We use the three dimensions to locate narrative components and additionally examine whether these distinctions correspond to separable SAE features. 


We use texts and their narrativity labels from the NarraDetect dataset \citep{piper2025narradetect} to identify components inside open-weight LMs whose activations correlate with narrativity. We ablate selected sets to assess their contributions to downstream tasks: text prediction, narrative generation, narrative inference, and hallucination. In doing so, our key findings are: 


\medskip

\begin{itemize}[noitemsep, topsep=0pt, leftmargin=*]
    \item \textbf{Identified component sets contribute to narrative representation.} Ablating narrativity-associated neurons and SAE features produces larger prediction-loss increases for narrative texts than for non-narrative texts.

    \item \textbf{Ablation of SAE narrative features reduces generation narrativity.} Ablation degrades generation broadly, but reduces the narrativity of generated text from Gemma and Llama with larger effects than size-matched random ablations that cannot be explained by changes in length or repetition alone.

    \item \textbf{There is some correspondence with narrative theory dimensions.} Existing feature explanations contain dimension-related information, and ablations show some preferential effects for agency and event sequencing, but not for world building. Explanations are also noisy and contain descriptions of grammar and syntax.

    \item \textbf{The consequences extend to some narrative-related benchmarks.} SAE ablation impairs story completion and narrative question answering more than random ablation, without comparable impairment on OpenBookQA. Multilingual effects are model-dependent. Effects on confabulation benchmarks are mixed.
\end{itemize}

\section{Related Work}





While classical narratology \citep{genette1980narrative} remains relevant and popular in story generation and understanding tasks \citep{liu2026narrative}, cognitive narratology \citep{herman2009basic,herman2013storytelling} examines beyond the text to study narrative as a cognitive resource. \citet{johnson2026characterizing} find that narrative dimensions which enable cognitive experientiality vary within pre-training sources. The notion of narrative as a cognitive resource is congruent with motivations behind StoryCloze \citep{mostafazadeh-etal-2016-corpus} and CosmosQA \citep{huang-etal-2019-cosmos}, framed as narrative benchmarks that test common-sense reasoning, causal inference, and contextual understanding. \citet{sui-etal-2024-confabulation} cite \citet{herman2013storytelling} when hypothesising LM confabulation/hallucination as types of narrative expression. 

\citet{piper2022toward} propose a data-driven theory of narrativity by operationalising formal dimensions from \citet{genette1980narrative} and \citet{herman2009basic} as linguistic features, combining them with lexical and syntactic \(n\)-grams, and applying statistical classifiers to identify textual patterns predictive of narrativity. Their work shows how narratology can use computational models to discover theories of narrative properties, but does not examine the internal mechanisms within computational models.

An LM's internal narrative representations have been studied using linear probes at different transformer layers to detect narrative incoherence \citep{de2026mary} and encoder model embeddings representing concepts such as characters \citep{volpetti2020temporal} and plot \citep{hatzel-biemann-2024-story}. Meanwhile, adjacent work in mechanistic interpretability has moved beyond detecting representations to intervening on model components associated with language variety, such as attention heads \citep{liu2026focusing}, neurons \citep{tang2024language}, and SAE features \citep{deng-etal-2025-unveiling}.

\section{Framework and Methodology}
\label{sec:method}

\subsection{Multidimensional Narrativity}
We measure narrativity as a scalar property (1--5 Likert, detailed definitions used are in Appendix Table \ref{app:narrativity-prompt}) using a three-part taxonomy which is \citet{piper2025narradetect}'s operationalisation of \citet{herman2009basic}'s cognitive narratology:
\[
\mathrm{Narrativity}
=
\operatorname{mean}(\mathrm{Agency},\,\mathrm{Event},\,\mathrm{World})
\]



\paragraph{Agency.}
``Narrative roots itself in the lived, felt experience of human or human-like agents interacting in an ongoing way with their surrounding environment'' \citep[p.~21]{herman2009basic}. \emph{Narrativity depends on the prominence of distinct agents who actively experience and respond to events.}

\paragraph{Event Sequencing.}
``Narrative... is a basic human strategy for coming to terms with time, process, and change'' \citep[p.~2]{herman2009basic}. \emph{Narrativity depends on the clarity and coherence with which event sequences are organised and presented.}

\paragraph{World Building.}
Stories function as blueprints for constructing ``storyworlds'' that readers can mentally inhabit and experience \citep{herman2009basic}. \emph{Narrativity depends on the extent to which a vivid and experienceable world is created, one that can be imagined, perceived, and felt by the reader.}

\subsection{Narrativity Value Correlation (NVC)}

\begin{figure}[h]
    \centering
    \includegraphics[width=\linewidth]{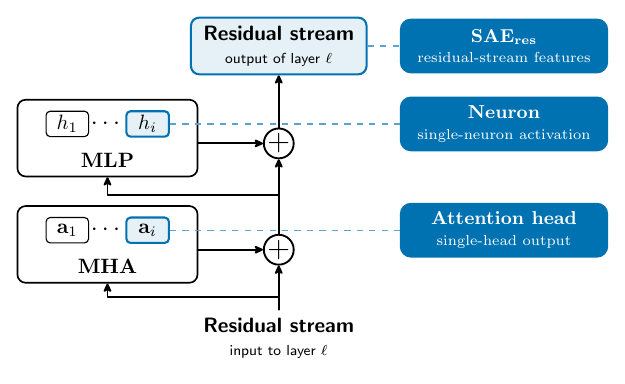}
    \caption{Finding NarraHeads, NarraNeurons and NarraFeatures in an LM's transformer layers.} 
    \label{fig:demo}
\end{figure}

We search for internal components at each transformer layer of LMs that show narrativity correlated behaviour. We refer to these component sets as \textit{NarraComponents}: \textit{NarraHeads}, \textit{NarraNeurons}, \textit{NarraFeatures} (Figure \ref{fig:demo}). We measure token-level component contributions as the L2 norm of attention head output vectors \citep{kobayashi-etal-2020-attention}, or as activation values for neurons and SAE features. For the latter, we attach pre-trained SAEs to each layer's residual stream. See Appendix \ref{appendix_maths} for additional implementation details.


For a working definition, (1) let $n_x$ denote the narrativity score of a given input text $x$, and $c_x$ the token-average contribution of a given component during the forward pass processing of text $x$. Furthermore, (2) let $\Delta \mathcal{L}_x$ denote the increase in the average next-token prediction loss during the forward pass processing of text $x$ when a component set's contribution is ablated. Given texts $A$ and $B$ where $n_A > n_B$, NarraComponents should generally satisfy both $c_A > c_B$ and $\Delta \mathcal{L}_A > \Delta \mathcal{L}_B$. We use Narrativity Value Correlation (NVC) to describe NarraComponents: their contribution values strongly correlate with the narrativity of the input, $\mathrm{NVC}_{\mathrm{cont}} = \operatorname{corr}(c, n)$; and the absence of their contribution (when ablated) causally affects the model's ability to represent texts (in next-token prediction) in a manner dependent on their degree of narrativity, $\mathrm{NVC}_{\mathrm{loss}} = \operatorname{corr}(\Delta \mathcal{L}, n)$.

We find NarraComponents by first recording component contributions when processing inputs with different narrativity (Section \ref{subsec:narradetect}), then computing Spearman's correlation between contribution value and narrativity ($\mathrm{NVC}_{\mathrm{cont}}$), and additionally between contribution value and each of the three dimensions $\mathrm{NVC}_{\mathrm{cont(a)}}$, $\mathrm{NVC}_{\mathrm{cont(e)}}$ and $\mathrm{NVC}_{\mathrm{cont(w)}}$. We choose Spearman because we make no assumption on the linearity of the relationship between narrativity and contribution, we only assume that it is monotonic. As a heuristic, we select components with strong correlations of $\rho \ge 0.5$ \citep{cohen2013statistical} (Figure \ref{fig:initial_selection}).




\subsection{SAE Feature Explanation Categorisation}
\label{sec:SAE_feature_explanation_categorisation}

For each NarraFeature identified from the pretrained SAEs, we retrieve its existing natural language explanation from Neuronpedia \citep{neuronpedia} (Section \ref{subsec:models}). We first categorise each feature  according to the narrativity dimension it correlates with most strongly. We use an embeddings classifier and zero-shot LLM to see whether those methods, using only the natural language explanations, can predict a feature's categorisation. We then test whether each subset's activations correlate with the narrativity dimensions of generated text, and ablate each subset separately to measure its on-target effect in open-ended generation (Section~\ref{sec:generation}).



\section{Experimental Setup}

\subsection{Narrative Dataset}
\label{subsec:narradetect}

We use the NarraDetect dataset \citep{piper2025narradetect}, which contains short passages (median word count: 104) from 18 genres. The `gold' human rated subset contains 394 items each with agency, event and world ratings in the range of 1--5. We first enrich the larger 13,530 NarraDetect dataset using LLM-generated `silver' ratings, which exhibit strong correlation with human ratings (Spearman's $\rho\ge0.795$; Appendix \ref{appendix_extend_narradetect}; Appendix Table \ref{tab:judge-human-spearman}), to initially identify NarraComponents and hold out the gold ratings for validation in Section \ref{sec:detection_verification}, ensuring that NarraComponents are identified and validated on different data.



\subsection{Models}
\label{subsec:models}

We use the pre-trained LMs GPT-2-Small-124M \citep{radford2019language}, Gemma-2-2B \citep{gemmateam2024gemma2} and Llama-3.1-8B \citep{team2024llama}, referred to here as GPT, Gemma and Llama; and the pre-trained SAEs GPT-2 autoencoders \citep{gao2025scaling}, Gemma Scope \citep{lieberum-etal-2024-gemma} and Llama Scope \citep{he2024llamascope}. We detail how we extract and ablate contributions in Appendix \ref{appendix_maths}.


Each targeted ablation is compared with a size-matched random baseline: the same number of components of the same type, sampled uniformly across all layers and averaged over 10 random sets (seeds 0–9) \citep{yamamoto2026neuron,wang2026heads}. Significance tests are two-sided paired bootstraps over prompts or benchmark items, Benjamini–Hochberg (BH) corrected within each table \citep{benjamini1995controlling}.

We retrieve the natural language SAE feature explanations from Neuronpedia \citep{neuronpedia}. These explanations were generated automatically using an unsupervised method \citep{bills2023language} prior to our experiment, independently and on different datasets \citep{gokaslan2019openwebtext,gao2020pile,soboleva2023slimpajama}. 


To predict the category of NarraFeatures, we compare (1) a supervised classifier that embeds each explanation using all-MiniLM-L6-v2 \citep{reimers-gurevych-2019-sentence} and train a three-class logistic-regression classifier (Appendix~\ref{app:topdim-classifier}); and (2) DeepSeek-V4.1-Flash \citep{xu2026deepseek}, prompted zero-shot to predict one of the same three categories.





\subsection{Generation}
\label{sec:generation_setup}

We build a bank of 900 prompts forming 450 pairs that differ only in the requested narrativity, e.g.\ \textit{``Write a \underline{story} about a teacher.''} vs.\ \textit{``Write an \underline{explanation} about a teacher.''} Pairs cover 90 topics (10 in each of nine categories, e.g.\ objects, emotions, conflicts, abstract concepts), each in five instruction templates (\textit{Write}, \textit{Give me}, \textit{Produce}, \textit{Provide}, \textit{Write \ldots\ concerning}) to reduce sensitivity to wording (Appendix Table~\ref{tab:opengen_topics}).

To reduce repetition and broken outputs from the base models, we use contrastive search \citep{su2022contrastive} in all of our generation experiments and detail the settings in Appendix \ref{app:generation}. We exclude GPT in the generation experiment as it fails to consistently produce coherent outputs.

We measure output narrativity using two frontier-class open-weight models accessed via API, DeepSeek-V4.1-Flash \citep{xu2026deepseek} and Hy3 \citep{tencent_hy3}, which exhibit strong correlation with human ratings on NarraDetect (Spearman's $\rho\ge0.83$; Appendix~\ref{app:llm_judge}; Appendix Table \ref{tab:judge-human-spearman}). In addition, we check a subset of outputs manually in Section \ref{sec:generation_ablation}.

\subsection{Benchmarks}

To test whether LM internal narrative representations contribute to the functional purposes described by \citet{herman2009basic}'s cognitive narratology, we measure the effects of ablation on six benchmarks across three categories:

\textbf{Narrative commonsense inference:}
XStoryCloze \citep{lin2022few} asks the model to select the coherent ending of a four-sentence story, requiring it to track agents and the causal and temporal development of events. Its extension of English StoryCloze \citep{mostafazadeh-etal-2016-corpus} to multiple languages lets us test whether features identified on the English-only NarraDetect transfer across languages; we evaluate English, Russian, Chinese and Hindi. CosmosQA \citep{huang-etal-2019-cosmos} uses personal blog narratives and asks about unstated causes, effects, facts and counterfactuals. These tasks test whether narrative representation serves as a resource for commonsense inference about what happened, why, and what may follow.

\textbf{Non-narrative QA (as control):}
OpenBookQA \citep{mihaylov-etal-2018-suit} asks elementary-science questions that require applying a supplied scientific fact to a novel situation. It tests general knowledge integration and multi-step reasoning without requiring the model to maintain agents, event sequences, or narrative coherence.

\textbf{Confabulation:}
\citet{sui-etal-2024-confabulation} argue that LLM hallucinations are \emph{confabulations}: narrative sense-making used to fill gaps in experience, and show that hallucinated outputs are more narrative than veridical ones. If narrative representations support this gap-filling process, ablating them should shift likelihood away from false or unsupported continuations. TruthfulQA \citep{lin-etal-2022-truthfulqa} tests preference for truthful answers over common misconceptions; FaithDial \citep{dziri-etal-2022-faithdial} contrasts faithful and hallucinated responses to supplied knowledge; and BEGIN \citep{dziri-etal-2022-evaluating} contrasts responses that are attributable or not attributable to provided evidence.

We follow the common benchmark convention of \citet{eval-harness} and evaluate accuracy on these benchmarks using log probability, detailed in Appendix \ref{app:benchmark-eval}. We exclude GPT in the benchmark experiments due to near-chance baseline performance and lack of multilingual capabilities.

\section{NarraDetect Results}
\label{sec_results}

\subsection{Detection and Verification}
\label{sec:detection_verification}

\begin{figure}[h]
    \centering
    \includegraphics[width=\linewidth]{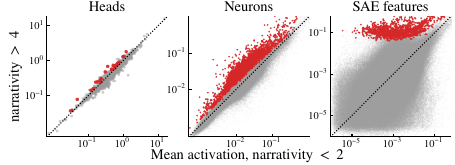}
    \caption{Mean activation of each component in Llama on low vs.\ high NarraDetect texts. Red marks NarraComponents selected by NVC. 
    }
    \label{fig:initial_selection}
\end{figure}


We find components in GPT, Gemma and Llama that meet the criterion of $\mathrm{NVC}_{\mathrm{cont}} \ge 0.5$, where NarraFeatures activate with stronger correlations than attention heads and neurons on narrative vs non-narrative inputs (Figure \ref{fig:initial_selection}). The correlation is weak in genres with low variation in narrativity, such as fables ($\rho$ between .10-.21), but reaches .34--.45 in historical texts, indicating that the components can be sensitive to narrativity beyond inter-genre differences (Appendix Table~\ref{tab:within-genre}).

\begin{figure}[h]
    \centering
    \includegraphics[width=\linewidth]{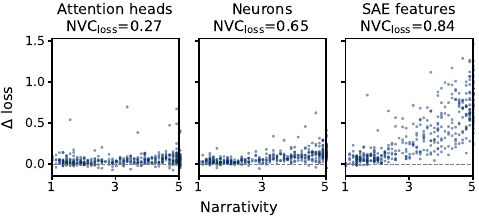}
    \caption{Change in loss correlates with narrativity when ablating NarraComponents inside Llama.} 
    \label{fig:ablation_loss}
\end{figure}

We ablate the identified components and observe changes to prediction loss (Table \ref{tab:selected-units}) on the human labelled texts from the NarraDetect gold dataset held out from detection. Ablating randomly selected components increases prediction loss similarly across texts with different narrativity but ablating NarraComponents identified using $\mathrm{NVC}_{\mathrm{cont}}$ yields selective effects that depend on text narrativity in 8/9 setups (Figure \ref{fig:ablation_loss}), with the only exception being the two identified GPT NarraHeads, which are the component type that has the weakest effects across all models (Table \ref{tab:selected-units}). 


\begin{table}[h]
\centering
\footnotesize
\setlength{\tabcolsep}{2.5pt}
\begin{tabular}{@{}l r r r@{\ }l r@{\ }l@{}}
\toprule
\textbf{Component} & \textbf{$n$} & \textbf{\%} & \multicolumn{2}{c}{\textbf{Narra}} & \multicolumn{2}{c}{\textbf{Random}} \\
\midrule
\multicolumn{7}{@{}l}{\textit{GPT}} \\
NarraHead    & 2 & 1.4 & $-$.01 & {\scriptsize[$-$.10,\,.10]} & .06 & {\scriptsize[$-$.01,\,.13]} \\
NarraNeuron  & 2534 & 6.9 & \textbf{.84} & {\scriptsize[.81,\,.87]} & $-$.04 & {\scriptsize[$-$.12,\,.04]} \\
NarraFeature & 695 & 0.2 & \textbf{.83} & {\scriptsize[.80,\,.86]} & .01 & {\scriptsize[$-$.08,\,.10]} \\
\addlinespace
\multicolumn{7}{@{}l}{\textit{Gemma}} \\
NarraHead    & 3 & 1.4 & \textbf{.13} & {\scriptsize[.03,\,.22]} & $-$.06 & {\scriptsize[$-$.15,\,.02]} \\
NarraNeuron  & 3470 & 1.4 & \textbf{.65} & {\scriptsize[.59,\,.71]} & .02 & {\scriptsize[$-$.04,\,.08]} \\
NarraFeature & 2529 & 0.6 & \textbf{.86} & {\scriptsize[.83,\,.88]} & .07 & {\scriptsize[$-$.06,\,.20]} \\
\addlinespace
\multicolumn{7}{@{}l}{\textit{Llama}} \\
NarraHead    & 28 & 2.7 & \textbf{.27} & {\scriptsize[.17,\,.37]} & $-$.06 & {\scriptsize[$-$.13,\,.02]} \\
NarraNeuron  & 4157 & 0.9 & \textbf{.65} & {\scriptsize[.58,\,.71]} & .01 & {\scriptsize[$-$.07,\,.09]} \\
NarraFeature & 1211 & 0.1 & \textbf{.84} & {\scriptsize[.81,\,.86]} & .04 & {\scriptsize[$-$.05,\,.14]} \\
\bottomrule
\end{tabular}
\caption{$\mathrm{NVC}_{\mathrm{loss}}$ when ablating NarraComponents vs.\ random: mean [95\% bootstrap CIs].}
\label{tab:selected-units}
\end{table}


\subsection{SAE NarraFeatures}


\begin{table}[h]
\centering
\footnotesize
\begin{tabular}{lrrr}
\toprule
\textbf{Category} & \textbf{GPT2} & \textbf{Gemma} & \textbf{Llama} \\
 & \scriptsize($n=695$) & \scriptsize($n=2529$) & \scriptsize($n=1211$) \\
\midrule
Agency               &  33.7\% &  40.2\% &  34.4\% \\
Event sequence       &  29.4\% &  21.7\% &  34.2\% \\
World building       &  37.0\% &  38.1\% &  31.5\% \\
\bottomrule
\end{tabular}
\caption{Percentage of NarraFeatures assigned to each dimension. 4435 in total.}
\label{tab:category-membership}
\end{table}


We assign each identified NarraFeature to the dimension whose score its activation correlates with most strongly (Table \ref{tab:category-membership}) and separately retrieve 4434/4435\footnote{One Llama feature has no explanation and is excluded from subsequent analysis but included during ablation.} explanation strings from Neuronpedia. We find examples of alignment: a Llama feature for ``expressions of personal growth and emotional experiences'' is most strongly correlated with agency ($\mathrm{NVC}_{\text{cont(a)}}{=}0.65$, vs.\ ${\le}0.52$ for the other two dimensions); a GPT feature for ``references to physical spaces and structural elements within a setting'' is tied to world building ($\mathrm{NVC}_{\text{cont(w)}}{=}0.64$, vs.\ ${\le}0.47$); a Gemma feature for ``sentences containing significant chronological or sequential events'' is tied to event sequencing ($\mathrm{NVC}_{\text{cont(e)}}{=}0.71$, vs.\ ${\le}0.53$).

\begin{table}[h]
\centering
\footnotesize
\setlength{\tabcolsep}{2.5pt}
\begin{tabular}{llrrrr}
\toprule
 & & \multicolumn{4}{c}{\textbf{\% of features}} \\
\cmidrule(l){3-6}
\textbf{Word} & \textbf{Top} & \textbf{All} & \textbf{Agency} & \textbf{Event} & \textbf{World} \\
 & & \scriptsize4434 & \scriptsize1666 & \scriptsize1167 & \scriptsize1601 \\
\midrule
action       & Event  & 23 & 15          & \textbf{33} & 24 \\
interaction  & World  & 16 & 11          & 18          & \textbf{19} \\
emotional    & Agency & 12 & \textbf{14} & 7           & 13 \\
character    & World  & 11 & 9           & 10          & \textbf{13} \\
personal     & Agency & 11 & \textbf{18} & 10          & 4 \\
narrative    & Event  & 10 & 7           & \textbf{14} & 10 \\
physical     & World  & 9  & 2           & 3           & \textbf{19} \\
relationship & Agency & 8  & \textbf{13} & 6           & 5 \\
experience   & Agency & 8  & \textbf{11} & 7           & 5 \\
event        & Event  & 7  & 4           & \textbf{15} & 5 \\
\bottomrule
\end{tabular}
\caption{The 10 most common words in the auto-generated explanations of NarraFeatures.}
\label{tab:topdim-keywords}
\end{table}

We also find that the most common words show narrative-related terms (Table~\ref{tab:topdim-keywords}): \textbf{Agency} contains ``personal'' and ``experience'', \textbf{Event} contains ``action'' and ``event'', and \textbf{World} contains ``physical''. This suggests that the auto-generated Neuronpedia explanations capture some meaningful distinctions between the three narrative dimensions. 

Many explanations describe grammar or syntax: ``punctuation and sentence endings'' (Gemma, $\mathrm{NVC}_{\text{cont}}{=}0.86$), ``punctuations and sentence-ending phrases'' (Llama, $0.81$), and ``punctuation marks, particularly periods and commas'' (GPT, $0.72$). Similar explanations can also end up with different dimension labels. Among Gemma features, ``punctuation marks and sentence endings'' falls under agency, ``end punctuation in sentences'' under event sequencing, and ``punctuation and sentence-ending cues'' under world building. 


\begin{table}[h]
\centering
\footnotesize
\setlength{\tabcolsep}{4pt}
\begin{tabular}{lrrr}
\toprule
 & \multicolumn{3}{c}{\textbf{Macro-F1}} \\
\cmidrule(l){2-4}
 & \textbf{GPT2} & \textbf{Gemma} & \textbf{Llama} \\
 & \scriptsize{n=695} & \scriptsize{n=2529} & \scriptsize{n=1210} \\
\midrule
Embedding Classifier & \textbf{.53} & \textbf{.57} & \textbf{.58} \\
DeepSeek Zero-shot & .50 & .47 & .48 \\
Chance & .33 & .33 & .33 \\
\bottomrule
\end{tabular}
\caption{Predicting a NarraFeature's top narrativity dimension from its auto-generated explanation.}
\label{tab:topdim-classifier}
\end{table}

We observe that classifiers are able to predict the categorisation above the .33 chance baseline (Table~\ref{tab:topdim-classifier}) and conclude that some narrativity-related signal exists in Neuronpedia’s explanation strings. We proceed by testing the causal effects of these dimension-specific subsets through ablating each category separately during open-ended story generation in Section~\ref{sec:generation_SAE_dimension}.

\section{Generation Results}
\label{sec:generation}

\subsection{Activations}

\begin{table}[h]
\centering
\footnotesize
\setlength{\tabcolsep}{2pt}
\setlength{\aboverulesep}{0pt}\setlength{\belowrulesep}{0pt}
\renewcommand{\arraystretch}{1.15}
\begin{tabular}{@{}l@{\hspace{3pt}}|@{\hspace{3pt}}rrr@{\hspace{3pt}}|@{\hspace{3pt}}rrr@{}}
\toprule
& \multicolumn{3}{c@{\hspace{3pt}}|@{\hspace{3pt}}}{Story} & \multicolumn{3}{c@{}}{Explanation} \\
& Agency & Event & World & Agency & Event & World \\
\midrule
\textit{Gemma} & \textbf{3.80} & \textbf{3.15} & \textbf{3.13} & 2.17 & 1.78 & 2.24 \\
\textit{Llama} & \textbf{4.05} & \textbf{3.27} & \textbf{3.28} & 1.78 & 1.49 & 1.96 \\
\bottomrule
\end{tabular}
\caption{Mean narrativity (1--5) of zero-shot generations, prompted for a story or an explanation.}
\label{tab:story_vs_explanation}
\end{table}

\begin{figure}[H]
    \centering
    \includegraphics[width=\linewidth]{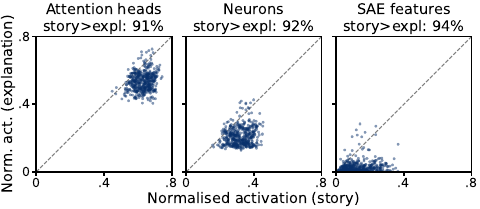}
    \caption{Llama's NarraComponents activate more for stories than explanations. Activations are max-normalised per component and averaged.
    }
    \label{fig:selection}
\end{figure}

When prompted to generate `story' vs `explanation', Gemma and Llama produce generations of different narrativity (Table \ref{tab:story_vs_explanation}). The previously identified NarraComponents activate more during story generation than explanation generation (Figure \ref{fig:selection}). We also find $\rho \geq 0.8$ correlations between our categorised NarraFeature contributions and their corresponding narrativity dimensional ratings in the generated outputs (Table \ref{tab:topdim_act_vs_score}).

\begin{table}[h]
\centering
\small
\setlength{\tabcolsep}{4pt}
\begin{tabular}{llccc}
\toprule
 & Features ($n$) & \multicolumn{3}{c}{Judge score} \\
\cmidrule(lr){3-5}
Model & by top dim. & Ag. & Ev. & Wo. \\
\midrule
Gemma-2-2B & agency (1016) & \textbf{0.82} & 0.74 & 0.72 \\
 & event (549) & 0.84 & \textbf{0.88} & 0.83 \\
 & world (964) & 0.76 & 0.81 & \textbf{0.89} \\
\midrule
Llama-3.1-8B & agency (416) & \textbf{0.82} & 0.73 & 0.70 \\
 & event (414) & 0.82 & \textbf{0.89} & 0.84 \\
 & world (381) & 0.73 & 0.80 & \textbf{0.88} \\
\bottomrule
\end{tabular}
\caption{Spearman $\rho$ between NarraFeature contributions and narrativity. 95\% bootstrap $\le\pm$0.04.}
\label{tab:topdim_act_vs_score}
\end{table}

\subsection{Ablation}
\label{sec:generation_ablation}

NarraFeature ablation changes output narrativity the most (Table \ref{tab:opengen-ablation-pooled}) and the effects are consistently and significantly larger than random ablation and sometimes results in Llama giving short replies, such as ``I don’t know what to write about a disappearance.'' The drop in narrativity remains significantly larger than under random ablation even after excluding prompts with degenerate or short outputs (Appendix Table~\ref{tab:degeneracy-sensitivity}).


\begin{table}[h]
\centering
\footnotesize
\setlength{\tabcolsep}{2pt}
\begin{tabular*}{\columnwidth}{@{\extracolsep{\fill}}lrrrrrr@{}}
\toprule
& \multicolumn{3}{c}{Target} & \multicolumn{3}{c@{}}{Random} \\
\cmidrule(lr){2-4}\cmidrule(l){5-7}
& Ag. & Ev. & Wo. & Ag. & Ev. & Wo. \\
\midrule
\multicolumn{7}{@{}l}{\textit{Gemma}} \\
NarraHeads & $-$.15 & $-$.17 & $-$.13 & $-$.13 & $-$.13 & $-$.12 \\[-2pt]
& {\scriptsize\color{gray}$\pm$.08} & {\scriptsize\color{gray}$\pm$.08} & {\scriptsize\color{gray}$\pm$.06} & {\scriptsize\color{gray}$\pm$.15} & {\scriptsize\color{gray}$\pm$.16} & {\scriptsize\color{gray}$\pm$.13} \\
NarraNeurons & \textbf{$-$.67} & \textbf{$-$.75} & \textbf{$-$.47} & $-$.11 & $-$.09 & $-$.08 \\[-2pt]
& {\scriptsize\color{gray}$\pm$.10} & {\scriptsize\color{gray}$\pm$.09} & {\scriptsize\color{gray}$\pm$.07} & {\scriptsize\color{gray}$\pm$.10} & {\scriptsize\color{gray}$\pm$.10} & {\scriptsize\color{gray}$\pm$.07} \\
NarraFeatures & \textbf{$-$1.86} & \textbf{$-$1.40} & \textbf{$-$1.27} & $-$.36 & $-$.44 & $-$.36 \\[-2pt]
& {\scriptsize\color{gray}$\pm$.11} & {\scriptsize\color{gray}$\pm$.10} & {\scriptsize\color{gray}$\pm$.08} & {\scriptsize\color{gray}$\pm$.16} & {\scriptsize\color{gray}$\pm$.15} & {\scriptsize\color{gray}$\pm$.12} \\
\addlinespace
\multicolumn{7}{@{}l}{\textit{Llama}} \\
NarraHeads & $-$.09 & $-$.20 & $-$.11 & $-$.07 & $-$.14 & $-$.10 \\[-2pt]
& {\scriptsize\color{gray}$\pm$.08} & {\scriptsize\color{gray}$\pm$.08} & {\scriptsize\color{gray}$\pm$.06} & {\scriptsize\color{gray}$\pm$.09} & {\scriptsize\color{gray}$\pm$.10} & {\scriptsize\color{gray}$\pm$.08} \\
NarraNeurons & \textbf{$-$.43} & \textbf{$-$.78} & \textbf{$-$.39} & $-$.04 & $-$.09 & $-$.06 \\[-2pt]
& {\scriptsize\color{gray}$\pm$.09} & {\scriptsize\color{gray}$\pm$.09} & {\scriptsize\color{gray}$\pm$.06} & {\scriptsize\color{gray}$\pm$.10} & {\scriptsize\color{gray}$\pm$.07} & {\scriptsize\color{gray}$\pm$.05} \\
NarraFeatures & \textbf{$-$1.53} & \textbf{$-$1.20} & \textbf{$-$1.03} & $-$.06 & $-$.11 & $-$.05 \\[-2pt]
& {\scriptsize\color{gray}$\pm$.10} & {\scriptsize\color{gray}$\pm$.09} & {\scriptsize\color{gray}$\pm$.07} & {\scriptsize\color{gray}$\pm$.07} & {\scriptsize\color{gray}$\pm$.08} & {\scriptsize\color{gray}$\pm$.07} \\
\bottomrule
\end{tabular*}
\caption{$\Delta Narrativity$ after ablating NarraComponents vs.\ random, $\pm$95\% CI. Bold: differs from random (BH $q<.05$).}
\label{tab:opengen-ablation-pooled}
\end{table}

Three authors acted as annotators and compared 90 pairs of Llama stories: NarraFeatures ablated vs random ablated ($\mathrm{seed}=0$). Each pair was shown unlabelled and shuffled in random order, and each annotator chose the more narrative-like text or declared a tie. Agreement was high (Krippendorff's ordinal $\alpha = 0.81$).\footnote{Ordinal $\alpha$ treats a disagreement between A and Tie as less than one between A and B; nominal $\alpha = 0.64$.} By majority vote, the annotators chose the random-ablation output as more narrative-like in 78 of the 90 pairs (87\%), indicating that narrative feature ablation predictably reduces a text's narrativity. We verified that the LLM judges gave the higher narrativity score to the authors' majority choice in 96\% (DeepSeek) and 97\% (Hy3) of pairs. 


\subsection{NarraFeature Dimensions}
\label{sec:generation_SAE_dimension}

\begin{table}[h]
\centering
\footnotesize
\setlength{\tabcolsep}{2pt}
\setlength{\aboverulesep}{0pt}\setlength{\belowrulesep}{0pt}
\renewcommand{\arraystretch}{1.15}
\begin{tabular}{@{}l@{\hspace{3pt}}|@{\hspace{3pt}}rrr@{\hspace{3pt}}|@{\hspace{3pt}}rrr@{}}
\toprule
& \multicolumn{3}{c@{\hspace{3pt}}|@{\hspace{3pt}}}{Story} & \multicolumn{3}{c@{}}{Explanation} \\
& Agency & Event & World & Agency & Event & World \\
\midrule
\textit{Gemma} & & & & & & \\
\quad agency & \textbf{$-$2.48} & $-$1.96 & $-$1.60 & \textbf{$-$1.09} & $-$0.70 & $-$0.75 \\
\quad event & $-$1.19 & \textbf{$-$1.55} & $-$0.92 & $-$0.30 & \textbf{$-$0.41} & $-$0.23 \\
\quad world & $-$1.10 & $-$1.17 & $-$0.77 & $-$0.22 & \textbf{$-$0.32} & $-$0.24 \\
\textit{Llama} & & & & & & \\
\quad agency & \textbf{$-$2.38} & $-$2.02 & $-$1.57 & \textbf{$-$0.61} & $-$0.30 & $-$0.39 \\
\quad event & $-$0.99 & \textbf{$-$1.59} & $-$0.91 & $-$0.10 & $-$0.16 & $-$0.05 \\
\quad world & $-$0.45 & \textbf{$-$0.64} & $-$0.44 & $-$0.07 & $-$0.06 & $-$0.03 \\
\bottomrule
\end{tabular}
\caption{$\Delta Narrativity$ after ablating each NarraFeature category. Bold: drop significantly larger than the next-largest drop in its row (BH $q<.05$). 95\% CIs $\le\pm0.15$.}
\label{tab:opengen-ablation-category}
\end{table}

\begin{table}[h]
\centering
\footnotesize
\setlength{\tabcolsep}{3pt}
\renewcommand{\arraystretch}{1.1}
\begin{tabular}{@{}l@{\hspace{4pt}}c@{\hspace{5pt}}p{0.66\columnwidth}@{}}
\toprule
\multicolumn{3}{@{}l@{}}{\textbf{Prompt:} \textit{Write a story concerning repairing a bicycle.}} \\
\midrule
Abl. & A/E/W & Generation (abridged) \\
\midrule
none & 5/5/4.5 & \textit{I was riding my bike to the store [\ldots] I saw a car coming towards me. I swerved [\ldots] hit a pothole and fell off my bike.} \\
\addlinespace[2pt]
agency & \textbf{1}/2.5/2 & \textit{Repairing a bicycle is a task that requires patience [\ldots] The first step in repairing a bicycle is to identify the problem.} \\
\addlinespace[2pt]
event & 4/\textbf{2}/3 & \textit{[\ldots] The boy's name is John. [\ldots] It has a flat tire. He can't ride it. He wants to ride it. He wants to go to the park. [\ldots]} \\
\addlinespace[2pt]
world & 5/5/\textbf{4} & \textit{I was riding my bicycle when I got a flat tire. [\ldots] I was very happy and I rode my bicycle home.} \\
\bottomrule
\end{tabular}
\caption{Llama generations after ablating each dimension's NarraFeatures (Agency/Event/World). More examples are in Appendix Table \ref{tab:opengen-random-examples}.}
\label{tab:opengen-ablation-example}
\end{table}

Ablating agency and event NarraFeature groups results in the largest decrease on the corresponding dimension. However, the interventions affect multiple dimensions, and agency-group ablation reduces event-sequencing scores more than event-group ablation does (Table \ref{tab:opengen-ablation-category}). We therefore interpret these results as partial alignment between the activation-based grouping and behavioural effects, rather than evidence for independently controllable dimensions. We do not observe a selective world-building effect. Table~\ref{tab:opengen-ablation-example} gives a qualitative example of on-target ablation effect.

\section{Benchmark Ablation Results}
\label{sec:benchmark}

\begin{table*}[h]
\centering
\footnotesize
\setlength{\tabcolsep}{2.5pt}
\resizebox{\textwidth}{!}{%
\begin{tabular}{lrrrrrrrrr}
\toprule
 & \multicolumn{4}{c}{\textit{Story completion (XStoryCloze)}} & \multicolumn{2}{c}{\textit{Question Answering (QA)}} & \multicolumn{3}{c}{\textit{Hallucination/Confabulation}} \\
\cmidrule(lr){2-5}\cmidrule(lr){6-7}\cmidrule(lr){8-10}
 & EN & RU & ZH & HI & Cosmos & OpenBook & TruthfulQA & FaithDial & BEGIN \\
\midrule
\multicolumn{10}{l}{\textit{Gemma-2-2B} (2529 features, 3 heads, 3470 neurons)} \\
Base & 82.7 & 74.7 & 73.3 & 63.7 & 42.6 & 69.4 & 25.0 & 67.6 & 85.5 \\
NarraFeatures & \textcolor[HTML]{B2182B}{\textbf{$-$9.1}}{\scriptsize$\pm$2.3} & \textcolor[HTML]{B2182B}{\textbf{$-$7.6}}{\scriptsize$\pm$2.3} & \textcolor[HTML]{B2182B}{\textbf{$-$5.0}}{\scriptsize$\pm$2.0} & \textcolor[HTML]{B2182B}{\textbf{$-$5.3}}{\scriptsize$\pm$1.7} & \textcolor[HTML]{B2182B}{\textbf{$-$2.8}}{\scriptsize$\pm$2.2} & \textcolor[HTML]{2166AC}{\textbf{+2.6}}{\scriptsize$\pm$2.5} & \textcolor[HTML]{2166AC}{\textbf{+2.8}}{\scriptsize$\pm$2.2} & \textcolor[HTML]{B2182B}{\textbf{$-$2.0}}{\scriptsize$\pm$1.2} & +0.5{\scriptsize$\pm$1.6} \\
Heads & $-$0.3{\scriptsize$\pm$1.0} & +0.7{\scriptsize$\pm$1.2} & $-$0.5{\scriptsize$\pm$1.0} & $-$0.5{\scriptsize$\pm$1.0} & $-$0.5{\scriptsize$\pm$1.4} & \textcolor[HTML]{2166AC}{\textbf{+2.3}}{\scriptsize$\pm$1.1} & $-$0.8{\scriptsize$\pm$1.0} & +0.3{\scriptsize$\pm$0.5} & $-$0.1{\scriptsize$\pm$0.7} \\
Neurons & $-$1.3{\scriptsize$\pm$1.2} & \textcolor[HTML]{B2182B}{\textbf{$-$2.2}}{\scriptsize$\pm$1.5} & $-$1.3{\scriptsize$\pm$1.3} & $-$0.9{\scriptsize$\pm$1.2} & +0.5{\scriptsize$\pm$1.7} & +0.4{\scriptsize$\pm$1.2} & +0.8{\scriptsize$\pm$1.3} & $-$0.2{\scriptsize$\pm$0.7} & +0.7{\scriptsize$\pm$0.8} \\
\midrule
\multicolumn{10}{l}{\textit{Llama-3.1-8B} (1211 features, 28 heads, 4157 neurons)} \\
Base & 85.0 & 76.2 & 76.4 & 68.0 & 81.2 & 86.9 & 28.9 & 68.0 & 85.8 \\
NarraFeatures & \textcolor[HTML]{B2182B}{\textbf{$-$6.4}}{\scriptsize$\pm$1.9} & $-$0.7{\scriptsize$\pm$1.7} & +0.2{\scriptsize$\pm$1.3} & $-$1.2{\scriptsize$\pm$1.4} & \textcolor[HTML]{B2182B}{\textbf{$-$1.6}}{\scriptsize$\pm$0.9} & $-$0.3{\scriptsize$\pm$0.4} & +2.5{\scriptsize$\pm$2.2} & $-$0.7{\scriptsize$\pm$1.1} & +0.5{\scriptsize$\pm$1.4} \\
Heads & \textcolor[HTML]{B2182B}{\textbf{$-$1.5}}{\scriptsize$\pm$1.0} & \textcolor[HTML]{B2182B}{\textbf{$-$2.8}}{\scriptsize$\pm$1.4} & \textcolor[HTML]{B2182B}{\textbf{$-$1.8}}{\scriptsize$\pm$1.2} & \textcolor[HTML]{B2182B}{\textbf{$-$2.2}}{\scriptsize$\pm$1.2} & +2.2{\scriptsize$\pm$2.1} & \textcolor[HTML]{2166AC}{\textbf{+1.3}}{\scriptsize$\pm$0.9} & 0.0{\scriptsize$\pm$1.3} & +0.3{\scriptsize$\pm$0.6} & +0.8{\scriptsize$\pm$0.9} \\
Neurons & +0.3{\scriptsize$\pm$1.0} & +0.3{\scriptsize$\pm$1.2} & +0.4{\scriptsize$\pm$1.1} & $-$0.6{\scriptsize$\pm$1.2} & \textcolor[HTML]{2166AC}{\textbf{+1.7}}{\scriptsize$\pm$1.6} & \textcolor[HTML]{2166AC}{\textbf{+1.7}}{\scriptsize$\pm$1.5} & +0.2{\scriptsize$\pm$1.7} & \textcolor[HTML]{B2182B}{\textbf{$-$1.2}}{\scriptsize$\pm$0.8} & $-$0.5{\scriptsize$\pm$0.6} \\
\bottomrule
\end{tabular}}
\caption{Effect of ablating NarraFeatures, NarraHeads and NarraNeurons relative to the same number of random units: $\Delta$Target$\,-\,\Delta$Rand (pp), $\pm$95\% CI. \textbf{Bold}: differs from random (BH $q<.05$; \textcolor[HTML]{B2182B}{red}: below, \textcolor[HTML]{2166AC}{blue}: above). Separate $\Delta$Target and $\Delta$Rand in Appendix Table~\ref{tab:heads_neurons_ablation}.}
\label{tab:sae_ablation_benchmarks}
\end{table*}


NarraFeatures ablations show the strongest and most consistent narrativity-dependent effects on next-token prediction loss and the largest reductions in output narrativity for Gemma and Llama, and we focus our benchmark analysis primarily on NarraFeatures but also report NarraHead and NarraNeuron ablations for completeness (Table \ref{tab:sae_ablation_benchmarks}).

Ablating NarraFeatures degrades XStoryCloze common-sense story completion significantly more than size-matched random ablation across all four languages tested on Gemma, but only for English on Llama. NarraHead and NarraNeuron effects on XStoryCloze are smaller than NarraFeature effects in English and on Gemma. However, on Llama, NarraHead ablation degrades non-English story completion more than both random and NarraFeature ablations do.

Ablation of NarraFeatures degrades the narrative CosmosQA performance more than size-matched random ablation. On the non-narrative OpenBookQA, degradation is similar to (Llama), or lesser than (Gemma), random ablation. These observations are consistent with our expectations of NarraFeatures. We find inconsistent ablation results in confabulation benchmarks (Table \ref{tab:sae_ablation_benchmarks}). 


\section{Discussion}

To our knowledge, this is the first work to use mechanistic interpretability to study storytelling and narrative theory. Our findings answer some questions about how LMs represent narrative, but they raise more than they settle. We therefore choose to focus the discussion on these open questions and the future directions they suggest.

\subsection{Narrative Components}


We identify NarraComponents among attention heads, FFN neurons, and SAE features. Under our selection (NVC) and intervention (zero ablation) procedure, ablating NarraFeatures produces the largest change in narrativity correlated prediction loss, open-ended generation narrativity, and degradation in English story completion. 

While present in Gemma, the lack of significant multilingual impact of NarraFeature ablation on Llama could reflect underlying differences in architecture or training. This opens up questions about \emph{the relationship between language varieties and internal LM narrative representation.}

Whether attention and FFN components support distinct aspects of narrative representation is also unanswered. Prior work interprets FFNs as key value memories that associate input patterns with output-relevant values \citep{geva-etal-2021-transformer}, whereas attention heads selectively move information between token positions \citep{vaswani2017attention,elhage2021mathematical} and mediate prediction-specific information-flow routes \citep{ferrando-voita-2024-information}. \emph{How do these different mechanisms interact to construct narrative representations?} 

The NVC method may be biased towards identifying certain types of features or components. The exploration of alternative methods, such as using entropy instead of correlation and probability instead of value \citep{tang2024language}, replacing ablation with mean or resample ablation
\citep{zhang2024towards}, or with steering that amplifies rather than removes
component contributions \citep{rimsky-etal-2024-steering}, is left for future work. Just as we identify high-NVC components (\(\rho \geq 0.5\)), we also observe components that are negatively correlated with narrativity (\(\rho \leq -0.5\)). Although these “anti-narrative” components fall outside the scope of our research questions, it is worth investigating \emph{how LMs represent ``negative narrativity''.}


\subsection{SAE Explanations}

We recognise recent criticism of SAEs, including evidence that learned feature decompositions may be non-unique and may not fully recover underlying monosemantic features \citep{paulo2026sparse,cui2026limits}, and that automatically generated explanations can be overly broad or fail to capture polysemanticity \citep{ma2025revising}. 

The Neuronpedia \citep{neuronpedia} explanations were generated before our study and on different datasets, but classifiers using only their text recover the activation-derived Agency, Event, and World categories above chance (Table~\ref{tab:topdim-classifier}). The explanations contain some meaningful dimension-specific signal, with their validity further supported by partial causal effects (Table \ref{tab:opengen-ablation-category} \& \ref{tab:opengen-ablation-example}).




\subsection{The Implicit Narrative Theory of the Language Model}

Originating in literary theory, narratology offers a set of theoretically grounded constructs for studying the logic and principles of narrative representation \citep{Meister2014Narratology}. To circumvent the infeasibility of studying the countless narratives of the world, classical narratology developed deductively by first imagining a hypothetical model of narrative, and then examining its compatibility with individual narratives \citep{barthes1975introduction}.
The combination of narratology and LM mechanistic interpretability opens new directions. LMs are artefacts that contain an inductively learned (drawing generalisations from examples), implicit theory of narrative. We propose LMs as unique objects of investigation for narratology.

Our findings offer preliminary insights into an LM's implicit narrative theory, suggesting a strongly agent-centred representation of narrative. Feature explanations emphasise action, interaction, emotion, characters, and relationships, while ablating agency features reduces all three narrativity dimensions and disrupts event sequencing more than ablating event features. World-building features exhibit the weakest and least selective effects. The representation also relies on surface linguistic cues with syntax and grammar. This interpretation remains tentative given the current limitations of SAEs.

Overall, it would seem that \citet{herman2009basic}'s definitions of narrative are at least partially compatible with some of an LM's internal representations, which the model learned without supervision. Repeating the experiment with another narrative theory, such as \citet{sternberg2011reconceptualizing}, may also yield findings to show \emph{which theories can and cannot describe LM narrative mechanisms.}

\subsection{Narrative for Intelligent Activity}

\citet{herman2013storytelling}'s cognitive narratology proposes the mind-narrative nexus: the interactions between stories and experiences. Applying this idea to LMs, the capacity for narrative may provide the necessary cognitive scaffolding for (artificially) intelligent activity: to represent the intentions, goals, emotions and conducts of self and other agents, the sequencing of events and causal relationships, and the construction of an experienceable world from semiotic representation. Our benchmark experiments follow from this premise and attempt to understand whether those representations contribute to common-sense inference, question answering, and confabulation.

Our positive results in XStoryCloze and CosmosQA suggest partial evidence for NarraFeatures being used for common-sense inference and reasoning. However, the assumption that sense-making can be tested through narrative related tasks leading to the conclusion that narrative representation is a resource for sense-making is circular reasoning. Future works need to explore how to test this in diverse and meaningful ways.

The mixed effects on the confabulation benchmarks may partly reflect that they capture hallucination in different ways. TruthfulQA evaluates preference for verified answers over human-curated misconceptions; FaithDial contrasts an original dialogue response with a human-edited version grounded in the supplied knowledge; and BEGIN compares model-generated responses according to whether they are attributable to the provided source. Narrative representations may consequently contribute to some forms of plausible gap-filling or confabulation without contributing equally to every form of falsehood or non-attribution. Moreover, our evaluation measures the relative log-probability assigned to fixed candidate responses. Further mechanistic work, such as evaluating ablated models on free-form generation, is needed to determine \emph{whether narrative representations causally contribute to confabulation.}


\subsection{Applications of narratology and mechanistic interpretability}

Narratology combined with mechanistic interpretability provides a new framework for investigating how LMs understand and generate narratives. Given documented limitations in LMs' narrative capabilities, including the homogeneity of generated narratives \cite{tian-etal-2024-large-language, hamilton2026elias}, difficulties maintaining narrative coherence \cite{li2026lost}, and weaknesses in reasoning about characters, events, and causal relationships \cite{yamin2024failure}, identifying the mechanisms underlying these behaviours (such as uneven distributions of narrative representations) may inform targeted interventions to improve narrative reasoning, understanding and generation. This has implications for domains in which narrative is central to social reasoning and communication, such as healthcare, psychotherapy, and the creative industries.

\section{Conclusion}

Through a framework which combines narratology and mechanistic interpretability, we identify attention heads, neurons and SAE features in GPT, Gemma and Llama that contribute to the models' internal representation of textual narrativity. We provide causal evidence to show that ablating components decreases output narrativity. We find that dimensions from \citet{herman2009basic}'s cognitive narratology show partial correspondence with the discovered SAE features inside the LMs. 


We test ablation effects on additional benchmarks. Despite being identified using an English-only dataset, NarraFeatures show cross-lingual effects in Gemma but not Llama. NarraFeature ablation also impairs narrative question answering on CosmosQA more than random ablation, but not on the non-narrative OpenBookQA benchmark. Results on confabulation benchmarks are mixed.

Overall, our findings provide evidence that LMs contain narrative mechanisms where certain components make causal contributions to model behaviour. The partial, dimension-specific, and model-dependent effects observed in our experiments also highlight substantial open questions about how narrative representations are organised, distributed, and used within language models. We propose these questions as new research directions with implications for theoretical narratology, LM interpretability, and practical applications which require narrative reasoning.

\section*{Limitations}

Our study is limited to three pre-trained language models. Components are selected using a heuristic threshold, and our causal claims apply to the ablated component sets collectively. Our random baselines are matched in size but not in layer distribution or contribution magnitude. The texts we study are mostly in English. We use pre-existing SAE feature explanations from Neuronpedia and we do not verify the process in which they were generated. We manually evaluate only a subset of outputs. Our random baselines are matched in size but not in layer distribution. 

\section*{Ethical Considerations}

The datasets include naturally occurring online text, and both
the datasets and pre-trained models may reproduce social, cultural, and
demographic biases present in their sources. Interventions that suppress or enhance
narrative representation may be useful for controlling output style, but could also
be used to manipulate rhetoric or alter framing without users' awareness.

\section*{Code and Data Availability}
Code will be made publicly available in a camera-ready version,
together with the datasets, and the lists of identified NarraComponents.

\section*{Declaration of AI Use}
We used generative AI for feedback and assistance during coding, analysis, and writing. All final
decisions were reviewed and made by the authors.

\bibliography{tacl2021}
\bibliographystyle{acl_natbib}


\onecolumn

\appendix

\section{LLM-as-Judge}
\label{app:llm_judge}

\begin{table}[H]
\centering
\footnotesize
\renewcommand{\arraystretch}{1.1}
\begin{tabularx}{\textwidth}{
    @{}
    >{\raggedright\arraybackslash}p{0.16\textwidth}
    >{\raggedright\arraybackslash}X
    @{}
}
\toprule
\textbf{Component} & \textbf{Prompt text} \\
\midrule

\textbf{System message:}
&
You are a professional literary analyst and expert in narratology.
\\
\midrule

\textbf{Instruction:}
&
Rate the following passage on three narrative criteria using
a 1-5 Likert scale. Base your scores on these specific definitions:
\\
\addlinespace[0.6em]

\textbf{1. Agency.}
&
Narrative is first and foremost a story about someone doing
something. Are there a limited number of agents who are clearly
foregrounded and experiencing something in this passage? I.e.
is there a protagonist or two primary characters that run through
all the actions? The more centralized and consistent one or more
agents are and the more coherent their identities, then the
stronger ``agency'' is as a quality. Jumping from one entity to
the next would constitute low agency, even if each of those
agents does something. How focalized is the passage around a
central figure(s)?
\\
\addlinespace[0.6em]

\textbf{2. Event Sequencing.}
&
Narrative is also about time and process. This occurs through
the sequencing of events. How clearly do you see a sequence of
events in your passage? I.e. can you put ``then'' into the
sentences easily (then this, then this)? They may be temporally
out of order (I tell what happened last first) or there may be
gaps between their happening (I walk into a restaurant and then
go home), and there may be simultaneity (something happens at
the same time as the thing before it), but there is an underlying
temporal relationship between the events that are mentioned in
the passage (i.e. they connect to each other in an experiential
way). While narrativity allows for digressions, reflections,
evaluations, etc., the more strongly structured a passage is
around event sequences, the higher this score should be.
Can these actions be experienced?
\\
\addlinespace[0.6em]

\textbf{3. World Making.}
&
As we are seeing, narratives are about lived experience.
Thus they require the construction of a world in which
sequenced actions undertaken by agents can transpire.
Think about this as ``inhabitability'' {-}{-} can you inhabit
what is happening in your ``mind's eye''? Can it be experienced?
Also key here is the unity of the world {-}{-} the actions all
make sense as part of a unified coherent space, even if this
is a fantastic space. Things are happening together. World
making isn't exclusively about description {-}{-} very
under-described worlds can be very concrete in terms of their
inhabitability. A fairy tale ``kingdom'' conjures a clear world
even if it is not particularly well described. Key is that
there is a ``thereness'' to the actions and events being invoked.
\\
\midrule

\textbf{Scale:}
&
1: Strongly disagree\newline
2: Somewhat disagree\newline
3: Unsure / Neutral\newline
4: Somewhat agree\newline
5: Strongly agree
\\
\addlinespace[0.6em]

\textbf{Output format}
&
Output ONLY the scores in this exact format, with NO explanation:
\newline
\texttt{Agency: [SCORE]}\newline
\texttt{Event: [SCORE]}\newline
\texttt{Worldmaking: [SCORE]}
\\
\addlinespace[0.6em]

\textbf{Passage}
&
\texttt{Passage: "\{passage\}"}
\\

\bottomrule
\end{tabularx}
\caption{Prompt used for LLM-as-judge labelling of Narrativity. Adapted from \citet{piper2025narradetect}.}
\label{app:narrativity-prompt}
\end{table}

To extend the NarraDetect corpus ($\sim$13k passages) in Appendix \ref{appendix_extend_narradetect}, we score each passage with local open-weight models: Gemma-4-31B \citep{team2026gemma}, GLM-4-32B \citep{glm2024chatglm}, and Qwen3.5-27B \citep{qwen35blog}. To evaluate open-ended generations from the ablated and unablated models (Section~\ref{sec:generation}), we use two frontier-class models accessed via OpenRouter API: DeepSeek-V4.1-Flash \citep{xu2026deepseek} and Hy3 \citep{tencent_hy3}. All models used to produce narrativity ratings show high alignment with human gold labels in NarraDetect (Table \ref{tab:judge-human-spearman}).

\section{Extending NarraDetect}
\label{appendix_extend_narradetect}


The NarraDetect dataset \citep{piper2025narradetect} contains five-sentence passages drawn from 18 genres: 11 narrative and 7 non-narrative genres (Appendix Table \ref{tab:within-genre}). The first, smaller sub-corpus is manually labelled: three trained annotators rated 394 passages (roughly 20 per genre) on a 1--5 scalar narrativity scale along the three dimensions described in Appendix Table \ref{app:narrativity-prompt}, with inter-rater agreement (average deviation index) of 0.41 for $\operatorname{mean}(\text{agency}, \text{event}, \text{world})$. We treat this 394-passage scalar subset as the gold-labelled set, and use LLM judges (Appendix \ref{app:llm_judge}) to extend narrativity ratings to the second larger unlabelled 13,530-document corpus (we remove 13 duplications from the original 13,543 dataset that matched texts in the gold 394 passages).  The three LLMs agree closely with one another, with inter-rater agreement (average deviation index) of 0.26 for the three-dimensional narrativity.



\section{Model Architecture}
\label{appendix_maths}
\subsection{Attention Heads}

Following \citet{kobayashi-etal-2020-attention}, we calculate an attention head's contribution by taking the output of each individual attention head and applying L2 vector norm. 

For notational simplicity, we consider a single attention head in a single layer and omit the corresponding head and layer indices. Let \(Q\), \(K\), and \(V\) denote the projected query, key, and value matrices. Its scaled dot-product attention output is

\begin{equation}
\begin{aligned}
\mathrm{head}
&= \operatorname{Attention}(Q,K,V) \\
&= \operatorname{softmax}\!\left(
    \frac{QK^{\top}}{\sqrt{d_k}}
    \right)V ,
\end{aligned}
\label{eq:attention-head}
\end{equation}

where \(d_k\) is the dimensionality of the key vectors and the softmax is applied over key positions. Before the multi-head output projection, the head output comprises one vector for each of the \(T\) token positions:

\begin{equation}
\mathrm{head}
=
\begin{bmatrix}
\mathrm{head}_1 \\
\mathrm{head}_2 \\
\vdots \\
\mathrm{head}_T
\end{bmatrix}
\in \mathbb{R}^{T \times d_v},
\qquad
\mathrm{head}_t \in \mathbb{R}^{d_v},
\label{eq:head-matrix}
\end{equation}

where \(d_v\) is the dimensionality of the value vectors and \(\mathrm{head}_t\) is the pre-output-projection vector produced for token position \(t\). 

For a given string of \(L\) tokens, we define the head's contribution value as the mean L2 norm of its token-level output vectors:

\begin{equation}
a
=
\frac{1}{L}
\sum_{t=1}^{L}
\left\lVert
\mathrm{head}_t
\right\rVert_2 .
\label{eq:head-activation}
\end{equation}

To measure the attention head's causal contribution, we ablate an individual attention head's contribution by setting its output value to an all-zero vector.

\subsection{Neurons}

\label{subsec:neruon_detection}



In a Transformer feed-forward network (FFN) module at layer $i$ \citep{vaswani2017attention}, the single-token representation is computed as:
\begin{equation}
    \bm{h}^{i} = \text{act\_fn}(\tilde{\bm{h}}^{i}\bm{W}_1^{i})\cdot\bm{W}_2^{i},
\end{equation}

where $\tilde{\bm{h}}^{i} \in \mathbb{R}^{d}$ is the input from residual stream after addition from the multi-head attention module, $\bm{W}_1^{i} \in \mathbb{R}^{d \times d_{\mathrm{ff}}}$, $\bm{W}_2^{i} \in \mathbb{R}^{d_{\mathrm{ff}} \times d}$, and $\text{act\_fn}(\cdot)$ denotes the activation function. Here, $d_{\mathrm{ff}}$ denotes the intermediate dimension of the FFN. We define the $j$-th neuron in the $i$-th FFN layer by the $j$-th column of $\bm{W}_1^{i}$, with its activation value given by $\text{act\_fn}(\tilde{\bm{h}}^{i}\bm{W}_1^{i})_j$. We consider a neuron's activation for a string as the average of all the individual token level activations.

To test whether the detected neurons have a causal effect, we ablate them. Let
$\bm{a}^{i} = \text{act\_fn}(\tilde{\bm{h}}^{i}\bm{W}_1^{i}) \in \mathbb{R}^{d_{\mathrm{ff}}}$
denote the neuron activations at layer $i$, and let $\mathcal{N}^{i}$ be the set
of detected neurons in that layer. We set the activation of each detected neuron
to zero, $a^{i}_j \leftarrow 0$ for all $j \in \mathcal{N}^{i}$, and leave all
other neurons unchanged. We apply the ablation at every token position (apart from the first attention sink position).




GPT-2 uses a plain MLP, $\mathrm{GELU}(xW_{\mathrm{in}})W_{\mathrm{out}}$, so we take $h = \mathrm{GELU}(xW_{\mathrm{in}})$.
Llama-3.1 and Gemma-2 use gated MLPs, $\big(\sigma(xW_{\mathrm{gate}}) \odot xW_{\mathrm{up}}\big)W_{\mathrm{down}}$, with $\sigma = \mathrm{SiLU}$ (SwiGLU) and $\mathrm{GELU}$ (GeGLU) respectively.
For these we take $h$ to be the input to $W_{\mathrm{down}}$.


\subsection{Sparse Autoencoders (SAEs)}

Individual model neurons may be polysemantic. 
Sparse autoencoders (SAEs) aim to address this problem by
representing a model activation as a sparse combination of learned
feature directions \citep{huben2024sparse}.

For notational simplicity, we consider one SAE at one Transformer
layer and omit the layer and token-position indices. Given a
residual-stream activation
$\mathbf{x}\in\mathbb{R}^{d}$, an SAE encodes it as

\begin{equation}
\mathbf{z}
=
\phi\left(
\mathbf{W}_{\mathrm{enc}}
(\mathbf{x}-\mathbf{b}_{\mathrm{dec}})
+
\mathbf{b}_{\mathrm{enc}}
\right),
\label{eq:sae-encoder}
\end{equation}

where
$\mathbf{W}_{\mathrm{enc}}\in\mathbb{R}^{m\times d}$ is the encoder
matrix, $\mathbf{b}_{\mathrm{enc}}$ and $\mathbf{b}_{\mathrm{dec}}$ are bias weights and $\mathbf{z}\in\mathbb{R}^{m}$ contains the resulting
feature activations. In the pretrained SAEs used in our experiments,
$m>d$, so the SAE maps the residual-stream activation into a
higher-dimensional feature space. The function $\phi$
makes this representation sparse by setting most entries of
$\mathbf{z}$ to zero.

The SAE decoder approximately reconstructs the original activation as

\begin{equation}
\hat{\mathbf{x}}
=
\mathbf{b}_{\mathrm{dec}}
+
\sum_{j=1}^{m}
z_j\mathbf{d}_j,
\label{eq:sae-decoder}
\end{equation}

where $z_j$ is the activation of feature $j$ and
$\mathbf{d}_j\in\mathbb{R}^{d}$ is its learned decoder direction.
Consequently, $z_j\mathbf{d}_j$ is the contribution of feature $j$ to
the SAE reconstruction.

We apply a layer-specific pre-trained SAE to the final residual stream
of every Transformer layer. For each input string, we average each
feature's activation across all token positions, producing one
activation value per feature, similarly to neurons from Appendix \ref{subsec:neruon_detection}.

To evaluate the causal effects of a selected set of features
$\mathcal{S}$, we subtract their decoder contributions directly from
the original residual-stream activation at each token position:

\begin{equation}
\mathbf{x}_{t}^{\mathrm{abl}}
=
\mathbf{x}_{t}
-
\sum_{j\in\mathcal{S}}
z_{t,j}\mathbf{d}_j.
\label{eq:sae-ablation}
\end{equation}


For each model we use the residual-stream SAEs at the output of every layer (12, 26 and 32 layers in total respectively for GPT, Gemma and Llama). The SAE dictionaries differ in size relative to their host models. The GPT-2 SAEs have 32k latents over a 768-dimensional residual stream, a $\approx$43$\times$ expansion. The Gemma Scope SAEs have 16k latents over 2,304 dimensions ($\approx$7$\times$), and for each layer we use the canonical SAE, defined in the Gemma Scope release as the one with average $L_0$ closest to 100 (59–141 across layers)\footnote{\url{https://huggingface.co/google/gemma-scope-2b-pt-res}}. The Llama Scope SAEs have 32k latents over 4,096 dimensions ($8\times$). The SAEs also use different sparsity mechanisms. The OpenAI GPT-2 SAEs use a TopK activation ($k=32$). The Llama Scope SAEs are trained with TopK and converted to a JumpReLU for inference. The Gemma Scope SAEs are trained and used with JumpReLU. 


Software versions: PyTorch~2.11.0, Transformers~5.14.1, SAELens~6.50.0.
Every SAE reads the residual stream at the output of a transformer block
(\texttt{blocks.$\ell$.hook\_resid\_post} in TransformerLens notation),
which we capture with a forward hook on the output of each decoder layer
of the Hugging Face implementation. 
Model weights and activations are in bfloat16. When computing correlations,
SAEs are encoded in bfloat16 with the SAELens forward pass. Cached feature
activations are stored in float16 and upcast to float32 for rank and
correlation statistics.


\section{Text Preprocessing and Token Handling}
\label{app:preprocessing}

Texts are tokenized with each model's own tokenizer, with its default special tokens and no other preprocessing. Gemma-2 and Llama-3.1 prepend a BOS token. GPT-2 has none, so we prepend \texttt{<|endoftext|>} as a BOS placeholder. Every sequence therefore begins with a special token at position~0. Sequences are truncated to the model's context limit. In practice this affects only 3 of 13,530 silver texts, all for GPT-2 (1,024 tokens). No gold text exceeds 630 tokens.

$\mathrm{NVC}$ correlations pool activations over content tokens only, excluding BOS and padding. Ablations do not modify the BOS position. BOS acts as an attention sink with outlier activations. Ablating at BOS would therefore mainly measure damage to the attention sink rather than narrativity-specific effects.


\section{Explanation Classifier Setup}
\label{app:topdim-classifier}

Inputs are the GPT-4o-mini Neuronpedia explanations (using the method of \citet{bills2023language}) of all labelled $n=4{,}434$ NarraFeatures, pooled over the three models (one Llama feature has no explanation and is excluded). Explanations are first embedded with \texttt{all-MiniLM-L6-v2} \citep{reimers-gurevych-2019-sentence}. A multinomial logistic regression ($L_2$, $C=1$, balanced class weights, solver=`L-BFGS'; untuned) is trained on the embeddings. We train the classifier on the explanations of two models and test it on the third, for each choice of held-out model, and report macro-F1. We use scikit-learn~1.9.0, sentence-transformers~6.0.1, CPU. RNG seeds are 0.

\section{Open-Ended Generation}
\label{app:generation}


\begin{table}[H]
\centering
\scriptsize
\setlength{\tabcolsep}{2pt}
\renewcommand{\arraystretch}{0.7}
\begin{tabular}{@{}p{0.19\textwidth}p{0.78\textwidth}@{}}
\toprule
\textbf{Topic type} & \textbf{Topics} \\
\midrule
\raggedright \textit{Entities \& agents} & \raggedright a teacher; a dog; an octopus; an alien; a dragon; a robot; a family; a team of explorers; a colony of ants; a mermaid \tabularnewline
\raggedright \textit{Objects} & \raggedright a key; a photograph; a clock; a letter; a map; a ring; a telescope; an ancient coin; a magic mirror; a time machine \tabularnewline
\raggedright \textit{Places \& environments} & \raggedright a kitchen; a library; a marketplace; a forest; a river; an abandoned station; a deep-sea trench; a distant planet; a floating city; an enchanted forest \tabularnewline
\raggedright \textit{Emotions \& mental states} & \raggedright joy; gratitude; relief; jealousy; loneliness; fear; curiosity; uncertainty; nostalgia; ambivalence \tabularnewline
\raggedright \textit{Relationships \& social dynamics} & \raggedright friendship; kinship; cooperation; trust between strangers; rivalry; authority; belonging; estrangement; dependence; exclusion from a group \tabularnewline
\raggedright \textit{Activities \& processes} & \raggedright gardening; learning a language; preparing a meal; repairing a bicycle; painting; building a bridge; migration; metamorphosis; erosion; brewing a magic potion \tabularnewline
\raggedright \textit{Events \& changes} & \raggedright an arrival; a departure; a disappearance; a reunion; a power outage; a discovery; a loss; retirement; an eclipse; the awakening of a dragon \tabularnewline
\raggedright \textit{Conflicts \& dilemmas} & \raggedright a disagreement over money; competing loyalties; a broken promise; a choice between safety and freedom; a dispute over a boundary; a choice between honesty and kindness; competition for water; a disagreement over a rule; a conflict between ambition and duty; a dispute over a magic artifact \tabularnewline
\raggedright \textit{Abstract concepts \& principles} & \raggedright justice; chance; freedom; responsibility; identity; beauty; truth; impermanence; fairness; infinity \tabularnewline
\bottomrule
\end{tabular}
\caption{The 90 topics of the open-ended generation prompts, ten per topic type.}
\label{tab:opengen_topics}
\end{table}

The 900 prompts cross 90 topics with five instruction templates and two conditions, \emph{story} and \emph{explanation}.  The ten template and condition combinations are \textit{Write/Give me/Produce/Provide \{a story $|$ an explanation\} about <topic>} and \textit{Write \{a story $|$ an explanation\} concerning <topic>}. Topics are noun phrases that include their own article (e.g.\ \textit{a teacher}). The story and explanation versions of a prompt therefore differ only in the requested genre, which yields 450 prompts per condition.

Base models have no chat template, so each prompt is wrapped in a zero-shot instruction frame. The exact string passed to the tokenizer is \texttt{Instruction: <prompt>\textbackslash nResponse:}. We decode with contrastive search \citep{su2022contrastive}, with $\alpha = 0.6$ and $k = 4$. A sequence ends when the model emits its EOS token or after 128 new tokens. Base models sometimes continue past their answer and begin a new, invented turn of the instruction frame (e.g.\ a finished story followed by \texttt{Response:} and an explanation of the same topic). We therefore cut each generation at the first occurrence of \texttt{Instruction:} or \texttt{Response:}. This cut applies to 15.2\% of Gemma and 5.3\% of Llama generations across the conditions in the ablation tables (14.1\% and 5.8\% for the unablated models).

\subsection{Human Annotation of Narrativity}

\label{app:human-annotation}

The annotated pairs come from the ablated Llama story generations: one with all 1,211 NarraFeatures ablated, and one with 1,211 randomly drawn SAE features ablated (the first of our ten random draws, seed $=0$). To keep the items independent, we used one prompt per topic, always the first template (\textit{Write a story about <topic>}), which gives one pair for each of the 90 topics (ten per topic type, Appendix Table~\ref{tab:opengen_topics}). No pair was excluded, including those with degenerate outputs.

Annotators saw the prompt, Text~A and Text~B in a spreadsheet while blind to the generation condition. The NarraFeature-ablated text was Text~A in exactly 45 of the 90 pairs, and the positions and the order of the pairs were shuffled once with a fixed seed ($0$), so all annotators saw the same sheet. Llama reproduces some corrupted punctuation from web text, also when unablated (e.g.\ \texttt{\#8230;} for an ellipsis, \texttt{don!t} for \textit{don't}). For display, we restored it with fixed rules applied identically to both conditions, which changed 26 of the 180 texts (16 NarraFeature, 10 random). Three authors annotated all 90 pairs independently. The instructions, verbatim, were:

\begin{quote}\footnotesize
For each item, read Text A and Text B (both are continuations of the prompt shown).\\
Choose which text is MORE NARRATIVE-LIKE (reads more like a story. It contains: events, characters, sequence, a world etc).\\
Use the dropdown in column E: A, B, or Tie (no clear difference).\\
In my experience, ties can be pretty common.

\smallskip
Column F: flag a text as INCOHERENT only if it is garbled, self-contradictory, or reads as nonsense. Poor style, quality and odd characters do not necessarily count as incoherent/nonsense.\\
You would need to make the decision on where the line is between repetition and incoherence/nonsense (there's a lot of repetition in the examples).\\
Some texts contain odd characters inside words (e.g.\ `!' or `\#') copied from the source text. Ignore these; do not treat them as errors.
\end{quote}
The options in column F were \textit{None}, \textit{A only}, \textit{B only} and \textit{Both}.

\section{Benchmark Evaluation}
\label{app:benchmark-eval}

All benchmarks in Table~\ref{tab:sae_ablation_benchmarks} and Appendix Table~\ref{tab:heads_neurons_ablation} are evaluated zero-shot by scoring candidate continuations with the model's own next-token log-probabilities. Following the convention of \citet{eval-harness}, we use either the per-token mean $\ell/T$ or the per-byte score $\ell/B(y)$, where $B(y)$ is the UTF-8 byte length of $y$. The normalisations used for each benchmark are listed below.

\paragraph{Story completion (XStoryCloze).}
We use the evaluation split of the English, Russian, Chinese and Hindi portions of XStoryCloze \citep{lin2022few} ($n=1{,}511$ stories per language). The prediction is the candidate ending with the higher token-normalized score $\ell/T$. Chance accuracy is 50\%.

\paragraph{Multiple-choice QA (CosmosQA, OpenBookQA).}
The prompt lists the passage (CosmosQA; \citealp{huang-etal-2019-cosmos}) or the supporting fact (OpenBookQA; \citealp{mihaylov-etal-2018-suit}), then the question and the lettered options, and ends with \texttt{Answer:}. We read the next-token log-probability of each option letter at the final position and predict the argmax. We evaluate on the CosmosQA validation set ($n=2{,}985$) and on all OpenBookQA splits pooled ($n=5{,}957$). Chance accuracy is 25\%.

\paragraph{Hallucination (TruthfulQA, FaithDial, BEGIN).}
All three use byte-normalized scores $\ell/B(y)$.
\begin{itemize}
  \item \textbf{TruthfulQA} (MC1; \citealp{lin-etal-2022-truthfulqa}): all $n=817$ questions of the validation split. Context: \texttt{Q:~<question>\textbackslash nA:}. The prediction is correct if the single true answer scores higher than every false answer. Chance accuracy is 22.6\%. 
  
  
  \item \textbf{FaithDial} \citep{dziri-etal-2022-faithdial}: we use the test-split ($n=2{,}924$ pairs). Each pair consists of a faithful response and the hallucinated response for the same context. A pair is correct if the faithful response scores higher than the hallucinated one.

  \item \textbf{BEGIN} \citep{dziri-etal-2022-evaluating}: We keep the $n=676$ items that have at least one fully attributable and at least one not fully attributable response. An item is scored as correct only if \emph{every} attributable response scores higher than \emph{every} non-attributable one.

\end{itemize}


  

\clearpage
\section{Complementary Results (Tables)}
\label{app:complimentary_results}

\begin{table}[H]
\centering
\footnotesize
\setlength{\tabcolsep}{5pt}
\begin{tabular}{lcccc}
\toprule
Judge & Overall & Agency & Event & World \\
\midrule
Hy3                 & .886 {\scriptsize[.867, .902]} & .831 {\scriptsize[.804, .853]} & .838 {\scriptsize[.803, .867]} & .855 {\scriptsize[.830, .877]} \\
DeepSeek-V4.1-Flash & .888 {\scriptsize[.868, .904]} & .850 {\scriptsize[.826, .870]} & .846 {\scriptsize[.816, .871]} & .869 {\scriptsize[.848, .888]} \\
Gemma-4-31B         & .898 {\scriptsize[.880, .913]} & .858 {\scriptsize[.837, .877]} & .840 {\scriptsize[.807, .868]} & .853 {\scriptsize[.831, .871]} \\
GLM-4-32B           & .883 {\scriptsize[.861, .901]} & .857 {\scriptsize[.834, .877]} & .845 {\scriptsize[.815, .869]} & .795 {\scriptsize[.761, .823]} \\
Qwen3.5-27B         & .882 {\scriptsize[.862, .897]} & .835 {\scriptsize[.805, .861]} & .868 {\scriptsize[.846, .886]} & .809 {\scriptsize[.776, .836]} \\
\bottomrule
\end{tabular}
\caption{Spearman $\rho$ between LLM-judge narrativity ratings and the mean of three human annotators ($n=394$ passages). Brackets give 95\% bootstrap CIs.}
\label{tab:judge-human-spearman}
\end{table}

\begin{table}[H]
\centering
\footnotesize
\setlength{\tabcolsep}{2pt}
\renewcommand{\arraystretch}{0.7}
\begin{tabular*}{\textwidth}{@{\extracolsep{\fill}}lcccrcccr@{}}
\toprule
& \multicolumn{4}{c}{Gemma} & \multicolumn{4}{c@{}}{Llama} \\
\cmidrule(lr){2-5}\cmidrule(l){6-9}
Filter & Agency & Event & World & $n$ & Agency & Event & World & $n$ \\
\midrule
Keep all & \textbf{$-$1.86}~{\scriptsize\color{gray}$-$.36} & \textbf{$-$1.40}~{\scriptsize\color{gray}$-$.44} & \textbf{$-$1.27}~{\scriptsize\color{gray}$-$.36} & 900 & \textbf{$-$1.53}~{\scriptsize\color{gray}$-$.06} & \textbf{$-$1.20}~{\scriptsize\color{gray}$-$.11} & \textbf{$-$1.03}~{\scriptsize\color{gray}$-$.05} & 900 \\
Rep.\ $\geq$.8 & \textbf{$-$1.92}~{\scriptsize\color{gray}$-$.37} & \textbf{$-$1.45}~{\scriptsize\color{gray}$-$.46} & \textbf{$-$1.31}~{\scriptsize\color{gray}$-$.36} & 736 & \textbf{$-$1.53}~{\scriptsize\color{gray}$-$.07} & \textbf{$-$1.29}~{\scriptsize\color{gray}$-$.13} & \textbf{$-$1.07}~{\scriptsize\color{gray}$-$.06} & 716 \\
Rep.\ $\geq$.6 & \textbf{$-$2.02}~{\scriptsize\color{gray}$-$.37} & \textbf{$-$1.55}~{\scriptsize\color{gray}$-$.45} & \textbf{$-$1.37}~{\scriptsize\color{gray}$-$.35} & 409 & \textbf{$-$1.50}~{\scriptsize\color{gray}$-$.08} & \textbf{$-$1.37}~{\scriptsize\color{gray}$-$.14} & \textbf{$-$1.11}~{\scriptsize\color{gray}$-$.06} & 496 \\
Rep.\ $\geq$.4 & \textbf{$-$2.09}~{\scriptsize\color{gray}$-$.31} & \textbf{$-$1.81}~{\scriptsize\color{gray}$-$.49} & \textbf{$-$1.45}~{\scriptsize\color{gray}$-$.33} & 137 & \textbf{$-$1.41}~{\scriptsize\color{gray}$-$.07} & \textbf{$-$1.40}~{\scriptsize\color{gray}$-$.16} & \textbf{$-$1.09}~{\scriptsize\color{gray}$-$.07} & 305 \\
Short & \textbf{$-$1.82}~{\scriptsize\color{gray}$-$.27} & \textbf{$-$1.36}~{\scriptsize\color{gray}$-$.40} & \textbf{$-$1.31}~{\scriptsize\color{gray}$-$.33} & 327 & \textbf{$-$1.48}~{\scriptsize\color{gray}$-$.01} & \textbf{$-$1.04}~{\scriptsize\color{gray}$-$.03} & \textbf{$-$.95}~{\scriptsize\color{gray}.00} & 341 \\
Short + rep.\ $\geq$.6 & \textbf{$-$2.00}~{\scriptsize\color{gray}$-$.28} & \textbf{$-$1.50}~{\scriptsize\color{gray}$-$.40} & \textbf{$-$1.34}~{\scriptsize\color{gray}$-$.27} & 160 & \textbf{$-$1.38}~{\scriptsize\color{gray}.01} & \textbf{$-$1.13}~{\scriptsize\color{gray}$-$.08} & \textbf{$-$.93}~{\scriptsize\color{gray}$-$.02} & 166 \\
\bottomrule
\end{tabular*}
\caption{$\Delta$Narrativity (1--5) after NarraFeature ablation, with size-matched random SAE ablation (mean of 10 seeds) in gray. Filters remove prompts with degenerate outputs at a seq-rep-4 threshold (Rep.), outputs of at most 20 words (Short), or either. A prompt is removed if any of its unablated, NarraFeature, or random-ablation outputs fails; $n$ prompts remain. Bold: differs from random (BH $q<.05$). All NarraFeature drops exceed those of every random seed.}
\label{tab:degeneracy-sensitivity}
\end{table}

\begin{table}[H]
\centering

\begingroup
\scriptsize
\setlength{\tabcolsep}{2pt}
\renewcommand{\arraystretch}{1.0}

\begin{tabularx}{\textwidth}{%
  @{}l@{\hspace{2pt}}c@{\hspace{3pt}}%
  >{\raggedright\arraybackslash}X@{}}
\toprule
Abl. & A/E/W & Generation (first 30 words) \\
\midrule

none & 4.5/4.5/4 &
I wrote a story about a man who lived in a desert. He had a well
that was his only source of water. One day, a man came to his
[\ldots] \\

agency & 2/3.5/3 &
Water is a precious resource that is essential for life.
Competition for water can arise due to various factors such as
population growth, climate change, and economic development.
In this [\ldots] \\

event & 5/4/4 &
I am going to write a story about a man who is competing for
water. The man is in a desert. He is thirsty. He is looking
for water. He [\ldots] \\

world & 3.5/3.5/3 &
I wrote a story about a man who is trying to get water for his
crops. He has to go to the river to get water, but there is a
[\ldots] \\

\midrule

none & 4.5/3.5/3 &
I was a caterpillar. I was a butterfly. I was a caterpillar again. \\

agency & 1/1/1 &
Metamorphosis is a process of change. Transformation is a process
of change. Transformation is a process of change. Transformation
is a process of change. Transformation is a process of change.
[\ldots] \\

event & 1/1/1 &
I don't know if this is what you're looking for, but here's a
story about metamorphosis. It's! \\

world & 4/1.5/2.5 &
I was a caterpillar once. I was a caterpillar for a long time.
I was a caterpillar for so long that I forgot what it was like
to be anything [\ldots] \\

\midrule

none & 5/4.5/4 &
I was walking down the street when I saw a river. It was a
beautiful river. The water was clear and the rocks were smooth.
I sat down on a [\ldots] \\

agency & 1/1/2.5 &
River is a place where people gather. River is a place where
people meet. River is a place where people share their stories.
River is a place where people are [\ldots] \\

event & 1.5/1/3 &
I am going to write a story about a river. It is a river that is
in the middle of a forest. The river is very deep"it is so deep
[\ldots] \\

world & 4/4.5/4 &
I wrote a story about a river. It was a! river that was very
polluted. The people in the town didn't care about the river.
They just threw their garbage [\ldots] \\

\bottomrule
\end{tabularx}
\par
\endgroup

\caption{Randomly sampled Llama story prompts (seed 0; prompt IDs
in bold), unablated and after category ablation.
A/E/W: agency/event/world scores, averaged over two LLM judges
(1--5).}
\label{tab:opengen-random-examples}
\end{table}

\begin{table*}[t]
\centering
\footnotesize
\setlength{\tabcolsep}{4pt}
\begin{tabular}{@{}lrcccccccc@{}}
\toprule
& & \multicolumn{2}{c}{Narrativity} & \multicolumn{2}{c}{GPT-2 small} & \multicolumn{2}{c}{Gemma-2-2B} & \multicolumn{2}{c}{Llama-3.1-8B} \\
\cmidrule(lr){3-4}\cmidrule(lr){5-6}\cmidrule(lr){7-8}\cmidrule(lr){9-10}
Genre & $n$ & Mean & SD & Neurons & SAE & Neurons & SAE & Neurons & SAE \\
\textit{Selected components} & & & & \textit{2,534} & \textit{695} & \textit{3,470} & \textit{2,529} & \textit{4,157} & \textit{1,211} \\
\midrule
Fables & 258 & 4.54 & 0.30 & .15 & .21 & .15 & .19 & .10 & .17 \\
Legal$^\dagger$ & 1,054 & 1.36 & 0.32 & .09 & .04 & .06 & .05 & .07 & .06 \\
ROCStories & 976 & 4.41 & 0.36 & .11 & .17 & .09 & .14 & .09 & .18 \\
Abstracts$^\dagger$ & 950 & 1.61 & 0.38 & .09 & .06 & .06 & .07 & .05 & .07 \\
\addlinespace[3pt]
Reddit stories & 971 & 4.07 & 0.57 & .22 & .27 & .20 & .26 & .20 & .30 \\
Contemporary novels & 776 & 3.92 & 0.59 & .22 & .29 & .20 & .27 & .19 & .29 \\
Short stories & 450 & 3.83 & 0.64 & .26 & .35 & .22 & .30 & .22 & .36 \\
US Supreme Court$^\dagger$ & 930 & 1.52 & 0.68 & .23 & .25 & .19 & .25 & .22 & .28 \\
19th-century novels & 998 & 3.76 & 0.68 & .28 & .38 & .25 & .34 & .18 & .35 \\
Flash fiction & 391 & 3.74 & 0.69 & .20 & .25 & .19 & .26 & .20 & .28 \\
Biography & 898 & 3.40 & 0.70 & .25 & .29 & .24 & .30 & .25 & .33 \\
Book reviews$^\dagger$ & 776 & 2.49 & 0.70 & .28 & .27 & .25 & .29 & .24 & .27 \\
Aphorisms$^\dagger$ & 461 & 1.99 & 0.72 & .36 & .25 & .30 & .27 & .27 & .23 \\
Fairy tales & 740 & 4.14 & 0.74 & .26 & .35 & .24 & .32 & .22 & .35 \\
Philosophy$^\dagger$ & 519 & 1.73 & 0.77 & .29 & .26 & .23 & .29 & .20 & .25 \\
Memoir & 935 & 3.55 & 0.79 & .31 & .40 & .29 & .38 & .30 & .42 \\
History & 979 & 3.17 & 0.82 & .38 & .43 & .34 & .42 & .34 & .45 \\
Literary studies$^\dagger$ & 468 & 1.94 & 0.83 & .37 & .34 & .34 & .36 & .33 & .35 \\
\midrule
All genres & 13,530 & & & .23 & .29 & .20 & .27 & .19 & .29 \\
All genres, random components & & & & .00 & .00 & .00 & .00 & .01 & .00 \\
\bottomrule
\end{tabular}
\caption{Within-genre Spearman $\rho$ between activation and narrativity on the silver set, median over each model's selected units. Genres are ordered by the within-genre SD of narrativity; $^\dagger$non-narrative. The median $\rho$ rises with the SD across genres (Spearman .64--.89), as expected under restricted range. All genres: one $\rho$ over all texts, with ranks centred within each genre. Random components: mean of 10 random sets of the same size.}
\label{tab:within-genre}
\end{table*}

\begin{table*}[h]
\centering
\footnotesize
\setlength{\tabcolsep}{2.5pt}
\resizebox{\textwidth}{!}{%
\begin{tabular}{lrrrrrrrrr}
\toprule
 & \multicolumn{4}{c}{\textit{Story completion (XStoryCloze)}} & \multicolumn{2}{c}{\textit{Question Answering (QA)}} & \multicolumn{3}{c}{\textit{Hallucination/Confabulation}} \\
\cmidrule(lr){2-5}\cmidrule(lr){6-7}\cmidrule(lr){8-10}
 & EN & RU & ZH & HI & Cosmos & OpenBook & TruthfulQA & FaithDial & BEGIN \\
\midrule
\multicolumn{10}{l}{\textit{Gemma-2-2B} (2529 features, 3 heads, 3470 neurons)} \\
Base & 82.7 & 74.7 & 73.3 & 63.7 & 42.6 & 69.4 & 25.0 & 67.6 & 85.5 \\
$\Delta$NarraFeatures & \textcolor[HTML]{B2182B}{\textbf{$-$11.6}}{\scriptsize$\pm$2.3} & \textcolor[HTML]{B2182B}{\textbf{$-$10.2}}{\scriptsize$\pm$2.2} & \textcolor[HTML]{B2182B}{\textbf{$-$6.1}}{\scriptsize$\pm$2.0} & \textcolor[HTML]{B2182B}{\textbf{$-$5.4}}{\scriptsize$\pm$1.9} & \textcolor[HTML]{B2182B}{\textbf{$-$6.2}}{\scriptsize$\pm$1.6} & \textcolor[HTML]{2166AC}{\textbf{$-$0.5}}{\scriptsize$\pm$0.8} & \textcolor[HTML]{2166AC}{\textbf{+3.4}}{\scriptsize$\pm$2.4} & \textcolor[HTML]{B2182B}{\textbf{$-$3.2}}{\scriptsize$\pm$1.3} & +0.1{\scriptsize$\pm$1.6} \\
\quad$\Delta$Rand & $-$2.6{\scriptsize$\pm$1.3} & $-$2.6{\scriptsize$\pm$1.6} & $-$1.0{\scriptsize$\pm$1.1} & $-$0.1{\scriptsize$\pm$1.0} & $-$3.4{\scriptsize$\pm$2.0} & $-$3.1{\scriptsize$\pm$2.4} & +0.6{\scriptsize$\pm$1.4} & $-$1.2{\scriptsize$\pm$0.9} & $-$0.4{\scriptsize$\pm$1.0} \\
\addlinespace[3pt]
$\Delta$Heads & $-$1.0{\scriptsize$\pm$0.9} & $-$0.5{\scriptsize$\pm$0.9} & $-$0.9{\scriptsize$\pm$0.9} & $-$0.4{\scriptsize$\pm$1.0} & $-$2.2{\scriptsize$\pm$0.9} & \textcolor[HTML]{2166AC}{\textbf{+0.3}}{\scriptsize$\pm$0.5} & $-$0.9{\scriptsize$\pm$1.0} & +0.6{\scriptsize$\pm$0.5} & $-$0.3{\scriptsize$\pm$0.7} \\
\quad$\Delta$Rand & $-$0.7{\scriptsize$\pm$0.9} & $-$1.1{\scriptsize$\pm$1.0} & $-$0.4{\scriptsize$\pm$0.9} & +0.1{\scriptsize$\pm$0.8} & $-$1.7{\scriptsize$\pm$1.3} & $-$2.0{\scriptsize$\pm$1.1} & $-$0.1{\scriptsize$\pm$0.8} & +0.3{\scriptsize$\pm$0.5} & $-$0.2{\scriptsize$\pm$0.5} \\
\addlinespace[3pt]
$\Delta$Neurons & $-$2.1{\scriptsize$\pm$1.3} & \textcolor[HTML]{B2182B}{\textbf{$-$3.2}}{\scriptsize$\pm$1.5} & $-$1.8{\scriptsize$\pm$1.4} & $-$0.7{\scriptsize$\pm$1.3} & $-$1.1{\scriptsize$\pm$1.2} & $-$1.5{\scriptsize$\pm$0.7} & +1.1{\scriptsize$\pm$1.4} & $-$0.1{\scriptsize$\pm$0.8} & +0.6{\scriptsize$\pm$0.8} \\
\quad$\Delta$Rand & $-$0.7{\scriptsize$\pm$0.6} & $-$1.0{\scriptsize$\pm$0.7} & $-$0.5{\scriptsize$\pm$0.7} & +0.3{\scriptsize$\pm$0.7} & $-$1.6{\scriptsize$\pm$1.6} & $-$1.8{\scriptsize$\pm$1.1} & +0.3{\scriptsize$\pm$0.9} & +0.1{\scriptsize$\pm$0.5} & $-$0.1{\scriptsize$\pm$0.4} \\
\midrule
\multicolumn{10}{l}{\textit{Llama-3.1-8B} (1211 features, 28 heads, 4157 neurons)} \\
Base & 85.0 & 76.2 & 76.4 & 68.0 & 81.2 & 86.9 & 28.9 & 68.0 & 85.8 \\
$\Delta$NarraFeatures & \textcolor[HTML]{B2182B}{\textbf{$-$7.1}}{\scriptsize$\pm$1.9} & $-$1.6{\scriptsize$\pm$1.7} & $-$0.1{\scriptsize$\pm$1.4} & $-$1.5{\scriptsize$\pm$1.5} & \textcolor[HTML]{B2182B}{\textbf{$-$2.0}}{\scriptsize$\pm$0.9} & $-$0.5{\scriptsize$\pm$0.5} & +2.9{\scriptsize$\pm$2.3} & $-$0.4{\scriptsize$\pm$1.1} & +0.3{\scriptsize$\pm$1.4} \\
\quad$\Delta$Rand & $-$0.7{\scriptsize$\pm$0.6} & $-$0.9{\scriptsize$\pm$0.9} & $-$0.2{\scriptsize$\pm$0.6} & $-$0.2{\scriptsize$\pm$0.7} & $-$0.4{\scriptsize$\pm$0.4} & $-$0.2{\scriptsize$\pm$0.2} & +0.5{\scriptsize$\pm$0.9} & +0.3{\scriptsize$\pm$0.4} & $-$0.2{\scriptsize$\pm$0.4} \\
\addlinespace[3pt]
$\Delta$Heads & \textcolor[HTML]{B2182B}{\textbf{$-$2.1}}{\scriptsize$\pm$1.0} & \textcolor[HTML]{B2182B}{\textbf{$-$4.0}}{\scriptsize$\pm$1.5} & \textcolor[HTML]{B2182B}{\textbf{$-$3.0}}{\scriptsize$\pm$1.3} & \textcolor[HTML]{B2182B}{\textbf{$-$2.8}}{\scriptsize$\pm$1.3} & $-$4.0{\scriptsize$\pm$1.1} & \textcolor[HTML]{2166AC}{\textbf{$-$1.5}}{\scriptsize$\pm$0.6} & 0.0{\scriptsize$\pm$1.5} & +0.6{\scriptsize$\pm$0.7} & +0.4{\scriptsize$\pm$0.9} \\
\quad$\Delta$Rand & $-$0.6{\scriptsize$\pm$0.7} & $-$1.2{\scriptsize$\pm$1.0} & $-$1.2{\scriptsize$\pm$0.9} & $-$0.6{\scriptsize$\pm$1.0} & $-$6.2{\scriptsize$\pm$2.0} & $-$2.8{\scriptsize$\pm$0.9} & 0.0{\scriptsize$\pm$1.1} & +0.3{\scriptsize$\pm$0.6} & $-$0.4{\scriptsize$\pm$0.7} \\
\addlinespace[3pt]
$\Delta$Neurons & 0.0{\scriptsize$\pm$1.1} & $-$0.5{\scriptsize$\pm$1.3} & $-$0.1{\scriptsize$\pm$1.2} & $-$1.1{\scriptsize$\pm$1.4} & \textcolor[HTML]{2166AC}{\textbf{$-$0.5}}{\scriptsize$\pm$0.8} & \textcolor[HTML]{2166AC}{\textbf{$-$0.2}}{\scriptsize$\pm$0.4} & +0.4{\scriptsize$\pm$1.7} & \textcolor[HTML]{B2182B}{\textbf{$-$1.0}}{\scriptsize$\pm$0.8} & $-$0.6{\scriptsize$\pm$0.7} \\
\quad$\Delta$Rand & $-$0.3{\scriptsize$\pm$0.5} & $-$0.8{\scriptsize$\pm$0.7} & $-$0.5{\scriptsize$\pm$0.6} & $-$0.4{\scriptsize$\pm$0.8} & $-$2.2{\scriptsize$\pm$1.5} & $-$1.9{\scriptsize$\pm$1.5} & +0.2{\scriptsize$\pm$0.9} & +0.3{\scriptsize$\pm$0.5} & 0.0{\scriptsize$\pm$0.5} \\
\bottomrule
\end{tabular}}
\caption{$\Delta$Target and $\Delta$Rand shown separately, as in Table~\ref{tab:sae_ablation_benchmarks}.}
\label{tab:heads_neurons_ablation}
\end{table*}

\end{document}